\documentclass[pdflatex,sn-mathphys-num]{sn-jnl}

\usepackage{graphicx}%
\usepackage{multirow}%
\usepackage{amsmath,amssymb,amsfonts}%
\usepackage{amsthm}%
\usepackage{mathrsfs}%
\usepackage[title]{appendix}%
\usepackage{xcolor}%
\usepackage{textcomp}%
\usepackage{manyfoot}%
\usepackage{booktabs}%
\usepackage{algorithm}%
\usepackage{algorithmicx}%
\usepackage{algpseudocode}%
\usepackage{listings}%

\begin{document}

\title[Article Title]{Configural processing as an optimized strategy for robust object recognition in neural networks}


\author*[1,2]{\fnm{Hojin} \sur{Jang}}\email{hojin4671@korea.ac.kr}

\author[2]{\fnm{Pawan} \sur{Sinha}}\email{psinha@mit.edu}

\author*[3]{\fnm{Xavier} \sur{Boix}}\email{xboix@fujitsu.com}

\affil[1]{\orgdiv{Department of Brain and Cognitive Engineering}, \orgname{Korea University}, \orgaddress{\city{Seoul}, \country{South Korea}}}

\affil[2]{\orgdiv{Department of Brain and Cognitive Sciences}, \orgname{Massachusetts Institute of Technology}, \orgaddress{\city{Cambridge}, \state{MA}, \country{The United States}}}

\affil[3]{\orgdiv{Artificial Intelligence Laboratory}, \orgname{Fujitsu Research of America}, \orgaddress{\city{Silicon Valley}, \state{CA}, \country{The United States}}}


\abstract{Configural processing, the perception of spatial relationships among an object's components, is crucial for object recognition. However, the teleology and underlying neurocomputational mechanisms of such processing are still elusive, notwithstanding decades of research. We hypothesized that processing objects via configural cues provides a more robust means to recognizing them relative to local featural cues. We evaluated this hypothesis by devising identification tasks with composite letter stimuli and comparing different neural network models trained with either only local or configural cues available. We found that configural cues yielded more robust performance to geometric transformations such as rotation or scaling. Furthermore, when both features were simultaneously available, configural cues were favored over local featural cues. Layerwise analysis revealed that the sensitivity to configural cues emerged later relative to local feature cues, possibly contributing to the robustness to pixel-level transformations. Notably, this configural processing occurred in a purely feedforward manner, without the need for recurrent computations. Our findings with letter stimuli were successfully extended to naturalistic face images. Thus, our study provides neurocomputational evidence that configural processing emerges in a naïve network based on task contingencies, and is beneficial for robust object processing under varying viewing conditions.}




\maketitle

\section{Introduction}\label{sec1}

Configural processing refers to the perception and interpretation of the spatial relationships and configurations among an object's components, facilitating integrated and holistic recognition. Early research underscores the importance of configural cues in object recognition, highlighting the brain's sensitivity to spatial arrangements. Biederman's Recognition-by-Components theory posits that object recognition is driven by identifying simple geometric components and their configurations \citep{biederman1987recognition}. Further studies reveal that configural processing extends to complex categorization, as evidenced by expert bird watchers and car enthusiasts who can discern subtle differences in similar species or models due to enhanced configural processing abilities \citep{tanaka1991object, gauthier2000expertise, gauthier2002unraveling}. This expertise-based perspective indicates that extensive experience with specific categories amplifies our capacity for holistic object processing.

Humans, among many object categories, are natural face experts. Configural processing is particularly crucial in face recognition, where faces, characterized by similar structures and subtle variations, serve as an ideal domain for studying this phenomenon. Research demonstrates that humans are highly sensitive to the spatial configurations of facial components \citep{young1987configurational, tanaka1993parts, maurer2002many}. Neurotypical individuals can discern subtle differences in facial features, such as interpupillary distance and philtrum length, even when other features remain identical \citep{le2001early}. This holistic ability suggests face specificity over other object categories \citep{farah1998special, goffaux2003spatial, goffaux2006faces}. Thus, the study of configural processing offers insights into our specialized perceptual skills and the foundations of expert visual recognition.

Despite extensive research, however, the functional benefits of configural processing are not fully understood. It might seem that focusing on individual features could offer more advantages for expert recognition systems, yet evidence indicates the opposite. What are the advantages of configural processing that lead experts to adopt this strategy over a piecemeal approach?

The present study introduces a novel perspective driven by teleological considerations; why might a visual system develop a configural processing strategy? We hypothesize that prioritizing configural cues over local feature processing could be an ecologically driven strategy, optimizing recognition under diverse viewing scenarios. A few psychological studies provide the motivation for this hypothesis, indicating the crucial role of configural processing in facilitating face recognition under demanding conditions. For instance, \citet{mckone2008configural} found that configural processing remained consistent across views, while part-based processing was view-sensitive, implying configural processing's adaptive function in robust face recognition amidst varying local image details. \citet{mckone2009holistic} further suggested that holistic processing is effective across a broad distance range. \citet{piepers2012review} proposed that if the holistic representation of a face is based not on the shape of its features but on a relational structure anchored from the features' center points, then alterations in the face’s second-order configuration would be minimal, thus offering stable cues for recognition. While appealing, these proposals currently lack robust computational validation.

Recent progress in deep learning offers a viable framework to test various hypotheses in cognitive science. Studies indicate that deep neural networks, specifically those trained for object recognition tasks, serve as the most advanced models of the visual system \citep{kriegeskorte2015deep, yamins2016goal}, accurately predicting visual cortical responses in human and macaque brains \citep{khaligh2014deep, yamins2014performance, gucclu2015deep, cichy2016comparison, long2018mid, jang2024improved}. Moreover, these models exhibit a substantial concordance with human face recognition behaviors \citep{hill2019deep, grossman2019convergent, higgins2021unsupervised, jozwik2022face} (see reviews \citep{o2018face, o2021face}), thus offering a variety of testable hypotheses for object and face recognition research.

However, it is essential to first validate that deep learning models are suitable for probing hypotheses related to configural processing. Some studies suggest these models might not possess human-like configural processing mechanisms, often exhibiting a bias towards local feature-based processing rather than global processing \citep{geirhos2018imagenet, baker2018deep}. Another study reports that deep neural networks struggle to capture configural cues in shape recognition tasks \citep{baker2020local, baker2022deep}. 

In this study, we investigated whether deep learning models could come to utilize configural cues for recognition based on the contingencies of task demands during training and, if so, whether these models prioritized configural processing over local processing under challenging viewing conditions. To this end, we created composite letter stimuli and compared various neural network models trained on tasks with either exclusively local or configural cues. We found that neural networks not only came to encode configural cues but also demonstrated enhanced robustness to geometric transformations, such as rotation and scaling, when employing these cues. Furthermore, when both feature types were concurrently available, the models favored configural cues over local ones. A layer-by-layer analysis demonstrated an increasing sensitivity to configural cues compared to local features in higher network layers, which might explain their resilience to pixel-level changes. Notably, this configural processing did not seem to require recurrent computations. The results obtained using letter stimuli were also successfully replicated with real-world face images. In summary, our research demonstrates experience dependent genesis of configural processing strategies, and shows how they might contribute towards achieving robust and reliable recognition capabilities across diverse viewing conditions.

\section{Results}\label{sec2}

\subsection{Deep neural networks can effectively capture configural cues in recognition}\label{subsec2}
To investigate the role of local and configural processing in recognition, it is crucial to separate one from the other. This study utilized the EMNIST dataset \citep{cohen2017emnist} to create composite patterns, where different letters represented individual features (\textbf{Fig. 1A}). In total, 9 letters were selected: B, D, L, M, N, P, R, T, and U. Subsequently, two distinct tasks were designed: one focusing on local featural processing ("local task") and the other concentrating on configural processing ("configural task"). In the local task, distinct classes were characterized by a unique set of letters, sharing the same configuration. Conversely, in the configural task, all classes used an identical set of letters, but varied in their configurations. A total of 24 classes were generated for each task (\textbf{Supplementary Fig. 1}).

\begin{figure}[h]
\centering
\includegraphics[width=1\textwidth]{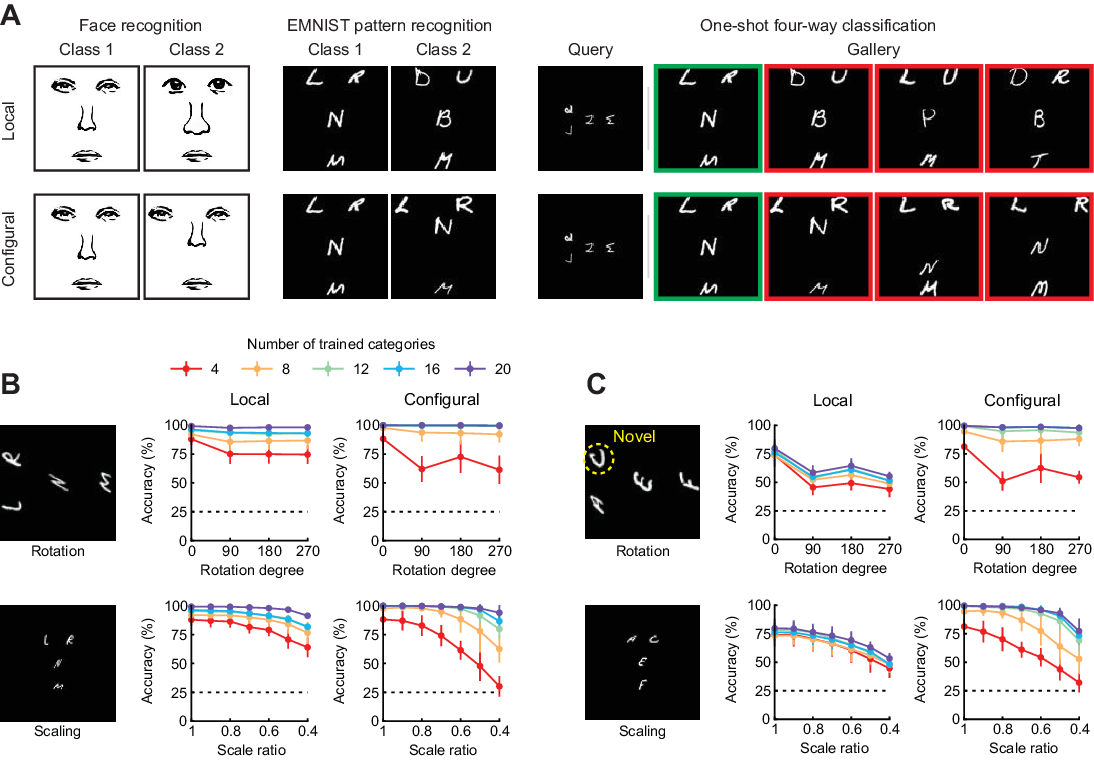}
\caption{\textbf{A} Conceptual illustrations (left) and actual representations (middle) of visual stimuli for local (top) and configural (bottom) tasks. Depiction of a one-shot four-way classification scenario with targets marked by green squares and distractors by red squares (right). \textbf{B} Performance accuracy for local and configural tasks under rotation (top) and scaling (bottom) transformations. The dashed line indicates chance level performance. Different colors represent the number of categories trained. \textbf{C} Performance accuracy for local and configural tasks under rotation (top) and scaling (bottom) transformations, with patterns that included novel local features.}\label{fig1}
\end{figure}

We evaluated two major types of neural network architectures, including feedforward and recurrent neural networks (detailed in the \textbf{Methods} section). To investigate the capability of these models in solving both tasks, we employed a one-shot four-way classification paradigm. Here, the networks were tasked with identifying a target class from four galleries, based on a query input from the same category as the target that underwent geometric transformations such as rotation and scaling (\textbf{Fig. 1A}). This one-shot learning paradigm allows for a more precise analysis of network generalization performance, particularly in how networks use local or configural cues to recognize novel classes. It also provides insights into how the diversity of training classes affects generalization \citep{jang2023robustness}. We varied the number of training classes to include 4, 8, 12, 16, and 20 out of a total of 24, and subsequently evaluated the network's ability to generalize to four patterns previously unseen by the system. Note that the total number of training images remained the same across all conditions.

\textbf{Fig. 1B} presents the accuracy levels for two tasks under two types of transformations: rotation (top) and scaling (bottom). The results revealed that the neural networks efficiently handled the local task, showing that local features were consistently effective for identification across various degrees of transformation. Consistent with the findings of \citet{jang2023robustness}, there was a noticeable improvement in generalization performance as the diversity of training classes expanded, even while the total amount of training images stayed the same. For the configural task, although initial performance was less than optimal with only four training classes, the networks quickly reached peak performance as the number of training classes increased, demonstrating effective use of configural cues. Collectively, these results underscore the efficacy of both configural and local feature cues within neural networks across different transformation conditions.

One could argue that our current methodological approach failed to clearly delineate local feature processing from configural processing. For instance, in the local task (\textbf{Supplementary Fig. 1}), networks could distinguish class 1 from class 2 based on the distinct letters ‘M’ and ‘T’. However, they might also rely on the spatial arrangement of ‘M’ with the other three letters to identify class 1, and similarly for ‘T’ to identify class 2. This highlights the challenge of separating local and configural cues in pattern recognition. Further analysis supports this, as networks trained on the local task demonstrated some ability to generalize to the configural task (\textbf{Supplementary Fig. 2A}). In contrast, those trained on the configural task failed in their generalization to the local task. To address this issue, one approach could be to randomly shuffle the locations of individual features during both training and evaluation phases. This strategy would ensure that the networks focus purely on the shape of local features without relying on their spatial relationships in the local task. By implementing this shuffling strategy, we were effectively able to differentiate local featural processing from configural processing, as the networks exhibited poor generalization between the local and configural tasks (\textbf{Supplementary Fig. 2B}). Furthermore, when the networks were tested under identical conditions to those used during training, they successfully addressed both local and configural tasks, reaffirming their capability to effectively utilize both local featural and configural cues.

\subsection{Configural processing is independent of individual local features}\label{subsec2}
Our previous results focused on the generalization capabilities of networks when presented with unseen letter patterns via a one-shot learning paradigm. We further escalated the challenge by assessing network performance on patterns composed of novel letters, specifically A, C, E, F, G, J, L, Q, and Y. Since these letters had not been previously encountered in a pattern by the networks, a decline in performance on the local task was anticipated. The primary objective of this investigation was to determine whether configural processing operates independently of local features or whether it is contingent upon a particular set of local features.

\textbf{Fig. 1C} shows the accuracy performance across both tasks when utilizing patterns comprising novel letters. As expected, the local task displayed a marked reduction in performance, though still surpassing the chance level. Importantly, the unaltered performance of the configural task, even with unfamiliar letters, highlights the independent nature of configural processing from local features. Similar results were replicated with the shuffling strategy (\textbf{Supplementary Fig. 3}). This result indicates the robust nature of configural cues in maintaining consistent recognition performance under varying viewing conditions, especially in scenarios where local features might be new or degraded.

\subsection{Configural cues are favored over local featural cues when both are concurrently available}\label{subsec2}
In real-world scenarios where a vision system has access to both local featural and configural information, which cue predominates? To answer this question, we introduced an additional task named the "local plus configural task" (as shown in \textbf{Fig. 2A}), where different classes were characterized by their unique local features and distinct feature configurations. Under these circumstances, the networks could choose to leverage either local features or configural information, likely favoring the more beneficial approach to maximize their performance. Subsequently, the networks were evaluated across the three tasks, i.e., the local, configural, and local plus configural tasks.

\begin{figure}[h]
\centering
\includegraphics[width=1\textwidth]{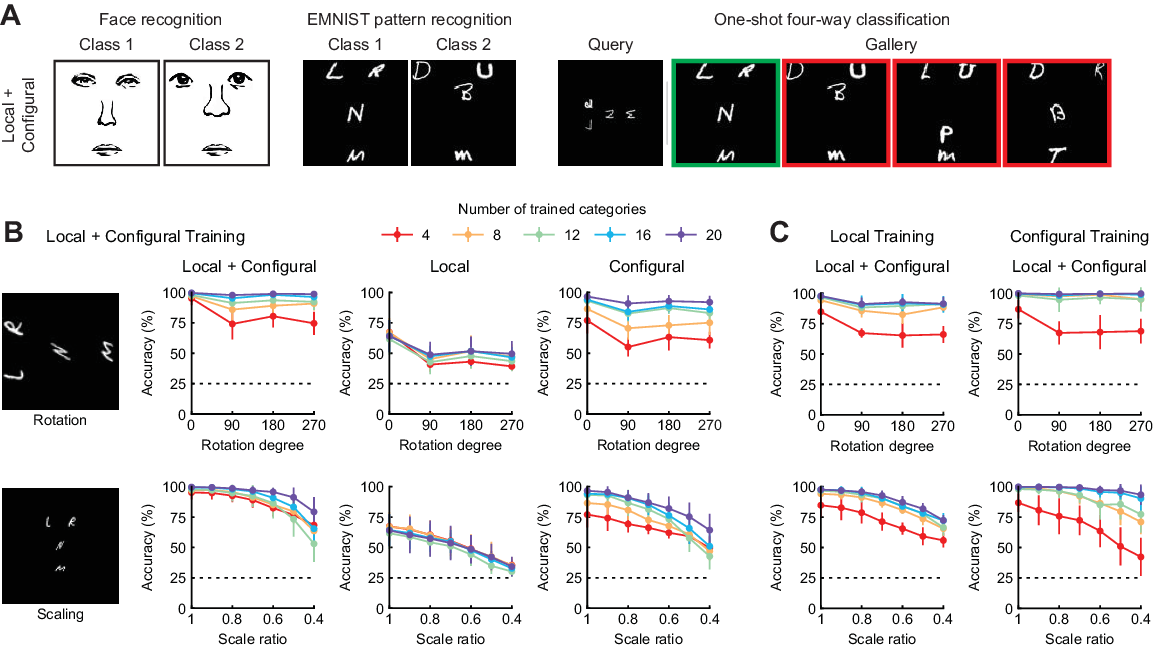}
\caption{\textbf{A} Illustration of the local plus configural task. \textbf{B} Performance accuracy of networks trained on the local plus configural task, when tested on the local plus configural task (left), the local task (middle), and the configural task (right), following the figure conventions described in \textbf{Fig. 1}. \textbf{C} Performance accuracy of networks trained on the local task and those trained on the configural task (left and right, respectively), each tested on the local plus configural task.}\label{fig2}
\end{figure}

\textbf{Fig. 2B} presents the networks' performance on the three tasks. When evaluated on the local plus configural task, networks exhibited high accuracy performance as expected. Notably, when the networks were evaluated independently on local and configural tasks, a pronounced superior performance in the configural task was observed. This finding highlights the networks' preference for configural cues, indicating that configural information may offer greater benefits for maintaining consistent performance under various transformation conditions. Additionally, in a reverse approach, networks trained on either the local or the configural task were subsequently assessed using the identical testing regime, namely the local plus configural task. While both networks effectively utilized each feature type, the network trained on the configural task demonstrated slightly more robust performance (\textbf{Fig. 2C}). Taken together, these observations provide evidence that configural information plays an important role in maintaining robust recognition performance across diverse conditions.

\begin{figure}[h]
\centering
\includegraphics[width=1\textwidth]{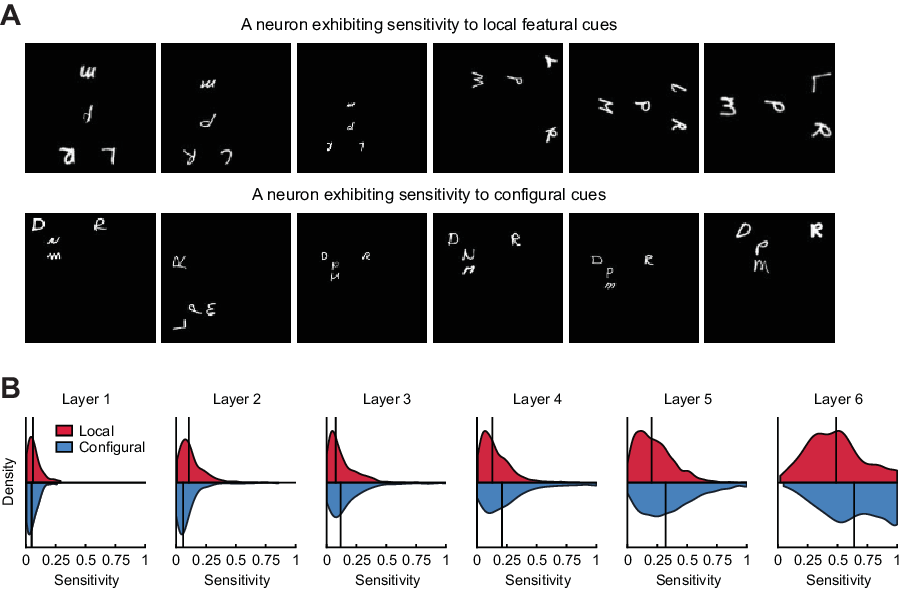}
\caption{\textbf{A} Top-6 images selected by a neuron sensitive to local featural cues (top) and another sensitive to configural cues (bottom). \textbf{B} Histograms displaying the sensitivity of individual neurons to local (red) and configural (blue) cues across the layers of ResNet50.}\label{fig3}
\end{figure}

To gain deeper insights, we performed an analysis at the single neuron level, examining the pattern of input images that elicited the strongest responses from individual neurons. Specifically, from the local plus configural task image set, we identified the top 20 images that maximally activated each neuron’s firing (\textbf{Fig. 3A}). By assessing the categorical consistency across these high-response images per neuron, we could determine whether that neuron exhibited selectivity for local featural or configural cues. This sensitivity was then quantified on a scale from 0 to 1 (details in the \textbf{Methods} section). \textbf{Fig. 3B} illustrates a histogram representing the distribution of neurons’ sensitivities towards local or configural cues across hierarchical layers. The result revealed that neurons in the early layers tended to be more tuned to local features, but this preference gradually transitioned toward configural cues in the later layers. This progression from local to configural cue tuning across layers may explain why configural processing exhibited greater robustness to pixel-level transformations.

\subsection{Impact of network architectures and training loss functions}\label{subsec2}
The local plus configural task's ability to reveal network biases toward local featural or configural cues could provide a framework for analyzing how different network architectures and training strategies influence a network's tendency to favor local or configural processing. Following the approach in \textbf{Fig. 2B}, networks trained on the local plus configural task, with varying architectures and loss functions, were assessed on either the local or the configural task. Here, instead of showing recognition performance across different rotation and scaling levels, we present the area under the performance curves as a function of the training classes.

\begin{figure}[h]
\centering
\includegraphics[width=1\textwidth]{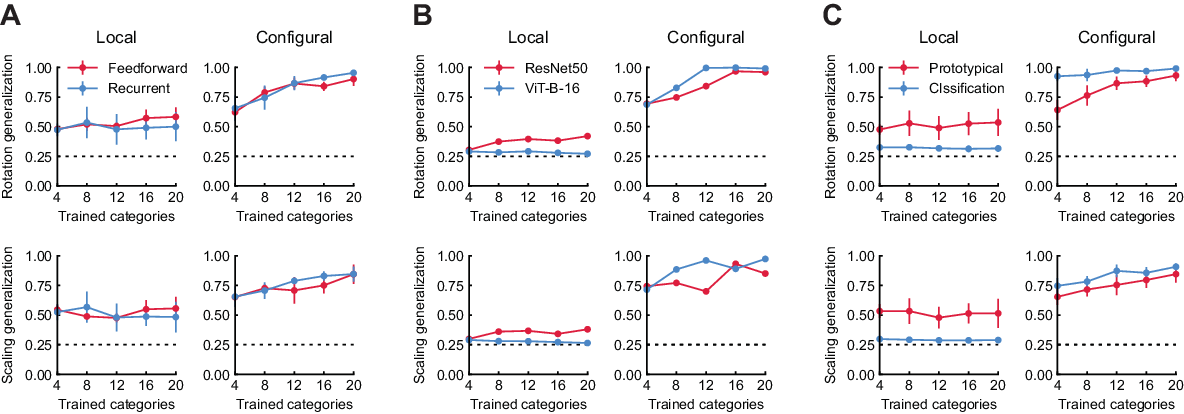}
\caption{Comparative analysis of generalization performance across rotation and scaling transformations among various neural network architectures and learning paradigms, including contrasts between feedforward and recurrent networks (\textbf{A}), ResNet50 versus ViT-B-16 (\textbf{B}), and a comparison of networks employing prototypical versus classification loss functions (\textbf{C}). All networks were trained on the local plus configural task and subsequently tested on the local (left) and configural (right) tasks, with the number of trained categories varying from 4 to 20.}\label{fig4}
\end{figure}

To begin, we investigated the potential impact of recurrent computations, which play an important role in robust object recognition \citep{wyatte2012limits, spoerer2017recurrent, kar2019evidence} and capturing long-range spatial dependencies \citep{sundaram2022recurrent}. \textbf{Fig. 4A} shows the comparison between feedforward and recurrent neural networks concerning their preferences for one type of the cues. The results suggest that both architectures leaned more toward configural cues, without noticeable differences between feedforward and recurrent network types. This observation suggests a minimal role for recurrent computations in enhancing configural processing.

Further analysis focused on the transformer architecture, assessing its bias towards configural rather than local featural cues. Current literature highlights the capability of the transformer architecture in capturing long-range dependencies through attentional modules \citep{dosovitskiy2020image, naseer2021intriguing, tuli2021convolutional, mao2022towards}, thus we hypothesized a stronger bias to configural cues than in conventional convolutional models. Given the data-intensive nature of training transformers, a comparative analysis between ImageNet pretrained Vision Transformer and ResNet was performed. Consistent with our hypothesis, the transformer architecture demonstrated a stronger bias towards configural cues over local featural ones relative to the convolutional network architecture (\textbf{Fig. 4B}).

Additionally, we evaluated the impact of different training loss functions on a network's featural processing strategy. By contrasting the prototypical loss function with the conventional classification loss function, we observed that networks employing the standard classification loss function exhibited a stronger reliance on configural cues (\textbf{Fig. 4C}). This finding implies that configural processing can be effectively modulated by manipulating the loss function employed during training.

\subsection{Generalization to real-world face stimuli}\label{subsec2}
Our findings with the EMNIST dataset led us to ask whether our results would generalize to face stimuli. Specifically, we sought to determine if neural networks, when trained on naturalistic facial stimuli such as those in FaceScrub \citep{ng2014data} under varying viewing conditions, i.e., rotation and scaling, would have a similar bias towards configural cues. To investigate this, we employed an open-source tool designed for crafting human avatar images \citep{bastioni2008ideas}. This allowed us to generate two facial sets: one with fixed configural elements and varying local ones and vice versa (see \textbf{Fig. 5A}). We then tested the performance of the FaceScrub-trained models on these face sets.

\begin{figure}[h]
\centering
\includegraphics[width=1\textwidth]{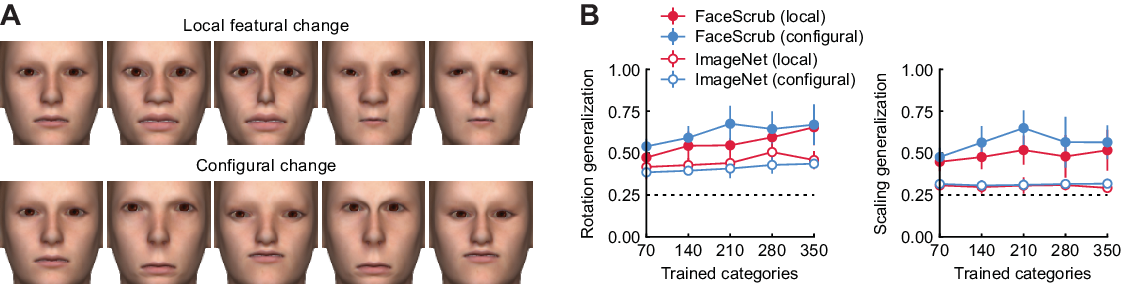}
\caption{\textbf{A} Examples of five facial stimuli are shown in two rows: top with differing local features but identical configurations, and bottom with identical local features but different configurations. \textbf{B} Generalization performance comparison across rotation (left) and scaling (right) transformations in networks trained on FaceScrub (filled circles) versus ImageNet (open circles), testing on facial stimuli that either vary in local features but share configurations (red) or share local features but vary in configuration (blue).}\label{fig5}
\end{figure}

As illustrated in \textbf{Fig. 5B}, a distinct pattern emerged between the two face sets. Under rotation or scaling transformations, the networks were notably better at recognizing faces with unique configural aspects over those with distinct local features. This supports our primary hypothesis about the significance of configural cues for robust face recognition. As an additional control experiment, we evaluated the same performance metrics but utilized networks trained on ImageNet. This was aimed at discerning if the bias to configural cues emerges intrinsically from exposure to facial stimuli, or is merely an artifact of the specific face stimuli used here, thus establishing a baseline. Our findings indicated that ImageNet-trained networks did not exhibit a preference for configural cues, further bolstering our original hypothesis. In summation, our computational evidence underscores the pivotal role of configural processing in recognition and the importance of training with classes that require configural processing.

\section{Discussion}\label{sec3}

In this study, we sought to test the hypothesis that configural processing may play a significant role in enhancing the robustness of visual recognition systems across varying viewing conditions. Although the existing literature offers some evidence supporting this hypothesis \citep{mckone2008configural, mckone2009holistic, jarudi2023recognizing}, such evidence lacks extensive validation through computational methods. As such, we leveraged deep neural network models to investigate the efficacy of configural cues in recognition by utilizing letter patterns subjected to geometric transformations, such as rotation and scaling.

Our results demonstrate that deep learning models effectively learn to discriminate between categories by leveraging configural cues, when these categories share the same local features (\textbf{Fig. 1B}). Moreover, our data reveal a specialized, independent mechanism for configural processing within these models, separate from the processing of local features (\textbf{Fig. 1C}), accentuating the critical role of configural cues in achieving reliable recognition across environments with varying local features. These findings demonstrate the ability of deep learning models to process configural cues, providing a counterpoint to findings from previous research \citep{baker2020local, baker2022deep}. Specifically, an earlier study reported that networks explicitly trained to classify based on global configurations while ignoring local elements still failed to capture global information, either exhibiting severe performance degradation upon local element distortion or showing no change in performance with fragmented global configurations \citep{baker2020local}. While making direct comparisons is difficult due to inherent differences across studies, the contrasting results could arise from several factors. Firstly, the stimulus sets employed in each study may have possessed distinct characteristics that influenced the models' ability to leverage configural cues. Secondly, variations in training regimens, such as the choice of loss functions, could have played a role, although we observed even more pronounced effects with the classification loss than with the prototypical loss. Thirdly, the extent of transformations applied to the stimuli, although involving similar operations like translation, rotation, and scaling in both studies, may have differed in their implementation details. Future studies are needed to clarify these contrasting findings.

The benefits of configural processing are more apparent when networks have access to both configural and local cues under varying viewing conditions. The preference for configural cues highlights their reliability for achieving optimal performance in such variable environments. This is in line with previous behavioral research, which has consistently shown that configural processing of faces is robust and effective, with changes in viewpoint and distance \citep{mckone2008configural, mckone2009holistic}. Our additional analyses with face stimuli, crafted to differentiate between local and configural cues, further confirm the consistent benefit of configural processing. Notably, this advantage of configural processing was observed exclusively in neural networks trained specifically for face recognition and was not present in networks trained for general object recognition. This suggests that the preference for configural cues in face recognition networks may evolve as an adaptive response to extensive and varied exposure to facial stimuli, rather than being an innate feature of the face recognition task. This observation proposes an intriguing hypothesis: holistic face processing may have developed as a strategic adaptation to ensure consistent and reliable recognition performance under diverse environmental conditions. This interpretation resonates with a line of research indicating that holistic processing results from extensive experience with stimuli \citep{diamond1986faces, gauthier1997becoming, gauthier2002unraveling}.

While our findings suggest that configural processing strategies may originate from extensive experience with an object under varying viewing conditions, we acknowledge other possibilities. For instance, clinical research presents a hypothesis that early limitations in visual acuity due to retinal immaturities might contribute to the development of configural face processing. Empirical evidence from individuals with congenital cataracts, who typically do not experience blurred vision initially, has revealed significant deficits in configural face processing in later stages of life \citep{le2001early, grand2004impairment}. Moreover, recent computational investigations have highlighted the potential developmental advantages of this initial blurred vision phase in enhancing integrative face processing \citep{vogelsang2018potential, jang2021convolutional}. While both hypotheses support the experience-dependent origin of configural processing strategies, future research should elucidate the roles of developmental factors and diverse viewing conditions to provide a comprehensive understanding of configural processing.

What underlies the superiority of configural processing relative to local processing within neural networks? To address this question, we conducted a detailed examination at the layer and unit levels of the network. We discovered that each unit was specialized to respond either to configural cues, local cues, or both, with this specialization varying by layer. Specifically, units in the lower layers were more responsive to local cues, while those in upper layers were more attuned to configural cues. This layer-specific responsiveness suggests a hierarchical processing approach, where local and configural cues are handled at different stages of visual perception. Neuroscientific evidence supports this hierarchy, indicating that the fusiform face area is principally involved in configural processing, whereas earlier cortical stages focus on local feature processing \citep{gauthier2002unraveling, yovel2005neural, schiltz2006faces, liu2010perception}. We propose that configural cues processed in later stages are inherently less affected by local pixel variations, thereby enhancing recognition stability across diverse conditions.

Are recurrent computations required for effective configural processing? While one might hypothesize that the integrative nature of recurrent computations is crucial for combining local features into holistic representations, our data do not support this view. We observed no significant difference in the processing of local versus configural information between recurrent and non-recurrent network architectures. This observation is corroborated by our layerwise analysis which demonstrates consistent unit preferences throughout the network layers, irrespective of recurrent cycles (\textbf{Supplementary Fig. 4}). Therefore, our results suggest that configural processing is likely driven by hierarchical computations rather than by recurrent dynamics, aligning with some studies suggesting that holistic processing may not depend on top-down attention \citep{boutet2002influence, norman2014spatial}.

Furthermore, our study indicates that the choice of network architectures and loss functions can impact a network's preference for local or configural cues. We showed that vision transformer models, known for their efficiency in capturing long-range dependencies \citep{dosovitskiy2020image, naseer2021intriguing, tuli2021convolutional, mao2022towards}, exhibited enhanced configural processing capabilities compared to convolutional neural networks. The transformer's self-attention mechanism allows for broad-scale integration of spatial cues, potentially facilitating more efficient configural visual information processing. Additionally, networks trained with prototypical loss functions, which focus on minimizing the distance to class prototypes, tend to favor local featural processing. Future investigations will be critical in elucidating the interplay between local and configural processing, thereby advancing the design of more robust and adaptable neural network models.

We also explored the effects of varying the number of training categories. An increase in category variety was found to improve the network's resilience to changes in rotation and scale, corroborating earlier research \citep{jang2023robustness} that underlines the benefits of diverse category experiences for network robustness. Unlike the previous work, however, the current method assessed network performance using entirely novel categories not previously exposed to transformations within a one-shot learning framework, thereby removing potential category selection bias associated with prior exposure to the transformations.

Lastly, this study contributes valuable insights into the intersecting domains of deep learning and psychological research, particularly in facial recognition. Despite extensive investigations into configural processing, computational analyses remain challenging due to limitations in existing frameworks and the difficulty of designing large-scale stimulus sets. Our research introduces a novel methodology, drawing significant inspiration from the rich history of psychological literature, that not only advances our understanding of the dynamics involved in facial configural processing but may also extend to practical aspects of face recognition techniques.

\section{Methods}\label{sec4}

\subsection{Visual stimulus set of letter patterns}\label{subsec4}
In this study, we constructed a novel set of visual stimuli composed of patterns. Each category included four letters arranged into a composite pattern. To introduce variability within a single letter, we collected five hundred distinct instances of each letter from the EMNIST database \citep{cohen2017emnist}. This diversity was crucial not only for providing a sufficient dataset for training neural network models but also for reflecting the natural variability in the appearance of individual features.

To systematically investigate the role of local and configural processing in recognition, we proposed two distinct tasks, as illustrated in \textbf{Fig. 1A}: the “local” and the “configural” tasks. In the local task, different categories shared identical configurations of individual letters but differed in their unique combinations of letters. We chose nine letters – B, D, L, M, N, P, R, T, and U – to generate a total of 24 distinct patterns. Conversely, in the configural task, while each category utilized an identical set of letters (L, R, N, and M), their configurations were unique. To prevent any overlap in the configurations, we segmented the entire input image (100 by 100 pixels) into 25 sections (20 by 20 pixels each) in a grid-like formation. Out of these, four were selected to create a total of 24 patterns. Details of the 24 categories are provided in \textbf{Supplementary Fig. 1}.

Additionally, to investigate whether configural processing is established on specific local features or is independent of them, we devised an additional stimulus set. This set was analogous to the original one but employed a different selection of 9 letters – A, C, E, F, G, J, L, Q, and Y – for the local task. Similarly, for the configural task, we utilized distinct configural patterns of the same 4 letters, namely A, C, E, and F. This new set of stimuli was only utilized for evaluation purposes and was not incorporated into the training process.

Beyond the local and configural tasks, we introduced an additional composite stimulus set termed the “local plus configural task”. This new task merged a category from the local task with one from the configural task, so that each category in this task presented both an exclusive set of local features and a distinct configuration. This experimental design was implemented to investigate which type of cue, local or configural, networks would prioritize when faced with both options simultaneously. Details and examples of the 24 categories are provided in \textbf{Supplementary Fig. 1}.

\subsection{Neural network architectures}\label{subsec4}
Our study conducted a comprehensive evaluation of various neural network architectures to verify the robustness and generalizability of our findings. This evaluation encompassed three feedforward convolutional neural networks: ResNet18, ResNet34, and ResNet50 \citep{he2016deep}. Concurrently, we analyzed four recurrent convolutional neural networks: CORnet-S \citep{kubilius2019brain}, BLnet, BLTnet \citep{spoerer2017recurrent}, and ConvLSTM \citep{shi2015convolutional}. For those recurrent models that incorporated batch normalization layers, we substituted these with layer normalization layers to mitigate potential batch effects. Each recurrent model underwent five recurrent cycles to allow for iterative integration of information. Detailed descriptions and methodologies of these models can be found in their respective original publications. Additionally, our investigation extended to vision transformers, specifically Vit-B-16 \citep{dosovitskiy2020image}. Due to the extensive training data requirements of this model, we employed a version of Vit-B-16 pretrained on the ImageNet dataset \citep{deng2009imagenet}. This model was compared with a similarly pretrained version of ResNet50 on ImageNet, to explore their relative performances in the local and configural tasks.

\subsection{Training procedures}\label{subsec4}
The networks underwent a two-stage training process using PyTorch, version 2.0.1. Initially, they were pre-trained on the classification of a single letter to develop robust low-level representations. This stage provided 2468 samples per letter, with the training images resized to 20x20 pixels and randomly positioned within a larger input field of 100x100 pixels. The training was conducted using the stochastic gradient descent optimization algorithm for 200 epochs, with a fixed learning rate of 0.001, a batch size of 64, a weight decay of 0.0001, and a momentum value of 0.9.

Following the initial pre-training, the networks were further optimized to recognize patterns of letters in one of the specific stimulus sets: the local, configural, or local plus configural tasks. In this subsequent training phase, the input patterns experienced a combination of transformations, including translation within the input field, rotation, and scaling. The rotation involved altering the orientation of the entire input pattern by 0, 90, 180, and 270 degrees, with all constituent elements of the pattern being rotated simultaneously. Scaling changed the size of the original 100x100 pixel input images, with ratios progressively reducing from a full scale to 0.4. 

This training phase adopted a four-way one-shot learning framework, utilizing the prototypical loss function proposed by \citet{snell2017prototypical}, which uses class prototype representations for query labeling. Training was conducted over 25,000 episodes using the stochastic gradient descent algorithm, with a fixed learning rate of 0.00001, a weight decay of 0.0001, and a momentum of 0.9. For each network, the final layer was modified to output 512 embedding vectors, which were then used to compute the prototypical loss. 

Additionally, the present study explored the impact of the number of training categories on network robustness, expanding on recent findings by \citet{jang2023robustness}. We systematically varied the number of training categories from 4 to 20, reserving the remaining four of 24 total categories for evaluation. Importantly, unlike the previous work, the current approach offered a legitimate assessment of invariance to transformations on novel stimuli, avoiding potential bias where invariant features might be associated with specific categories. 

\subsection{Layerwise neural sensitivity analysis}\label{subsec4}
In this analysis, the goal was to assess the sensitivity of individual neurons to local and configural cues. To achieve this, we selected the 20 images that elicited the strongest response from each neuron, using stimuli from the local plus configural task. We then evaluated the neuron's sensitivity to either local or configural cues by examining the consistency of categorical selections among these 20 images. This was accomplished by calculating the entropy, which quantifies the uncertainty or randomness of the categorical selections, as detailed below:
\begin{equation}
{S(X)} = 1 - {H(X)} / {\log_2N},
\end{equation}
where \(H(X)=-\sum\nolimits_{i=1}^{N}p(x_i)\log_2p(x_i)\) represents the entropy and N was the number of categories, which was 4 in this case. The sensitivity measure ranged from 0 to 1. A value of 1 signified maximum sensitivity, where the neuron consistently identified all 20 top images as belonging to a single category. This indicated that the neuron was highly specialized or tuned to the local or configural cues characterizing that specific category. Conversely, a value of 0 meant that the neuron exhibited complete insensitivity, with its selections from the 20 top images showing no discernible pattern or preference for any specific category, suggesting a lack of sensitivity to either local or configural cues.

\subsection{Visual stimulus set of faces}\label{subsec4}
We sought to test whether the configural processing bias observed with the EMNIST dataset could be also applicable to facial recognition tasks using natural face images. To this end, we trained neural networks on the FaceScrub database \citep{ng2014data}, which included facial images altered through rotation and scaling, following the same training procedures as those used for EMNIST but using a one-shot five-way learning paradigm instead. The number of identities varied systematically across training sets, including 70, 140, 210, 280, and 350. Each network model was trained using a fixed learning rate of 0.00001 through stochastic gradient descent, incorporating a batch size of 64, weight decay of 0.0001, and momentum of 0.9. Additionally, networks were also trained on ImageNet, using 350 object categories selected at random from a set of 1000, to serve as a baseline for comparison.

To determine the local or configural processing bias of neural networks trained with real-world face images, we utilized MakeHuman \citep{bastioni2008ideas} (http://www.makehumancommunity.org/), a tool for crafting human avatars, to generate specific sets of facial stimuli. Similar to the original approach, we developed two distinct types of stimuli: local and configural. For local feature recognition, we created faces with the same overall configuration but altered specific local features. More specifically, starting with an average face \citep{hays2020faret}, modifications were made in aspects such as the scaling of right and left eyes, positioning of the eye corners, and mouth scaling both horizontally and vertically, ended up with four distinct facial stimuli (as seen in the top panel of \textbf{Fig. 5A}). Conversely, for the configural task, while keeping local features constant, we varied the overall configuration of facial elements. This included adjustments in the positions of both eyes (inward/outward and upward/downward), along with changes in the nose and mouth configurations (shown in the bottom panel of \textbf{Fig. 5A}). To introduce more variations in the facial images for each condition, we altered the yaws (from -5 to 4 degrees) and pitches (from -5 to 4 degrees), and adjusted the lighting conditions with three different light sources along the x, y, and z axes. These sets of stimuli were used in evaluations of both face- and ImageNet-trained networks.

\clearpage\bibliography{sn-bibliography}

\clearpage\section*{Data Availability}
All stimulus sets employed in this study will be available on the Open Science Framework at https://osf.io/htduf/.

\section*{Code Availability}
The codes for training and analysis will be made available on the Open Science Framework at https://osf.io/htduf/. 

\section*{Acknowledgements}
This research was supported partly by the following grants from Korea University, K2404751, and also by the Office of the Director of National Intelligence (ODNI), Intelligence Advanced Research Projects Activity (IARPA), via 2022-21102100009. The views and conclusions contained herein are those of the authors and should not be interpreted as necessarily representing the official policies, either expressed or implied, of ODNI, IARPA, or the U.S. Government. The U.S. Government is authorized to reproduce and distribute reprints for governmental purposes notwithstanding any copyright annotation therein. 

\section*{Author Contributions}
H.J. and X.B. conceived of and designed the study, H.J. trained neural network models and performed all data analyses, H.J., P.S., and X.B. wrote the manuscript together.

\section*{Competing Interests}
The authors have no competing interests.

\clearpage\begin{appendices}
\renewcommand{\figurename}{Supplementary Fig.}
\setcounter{figure}{0}    

\section*{Supplementary Information}

\begin{figure}[h]
\centering
\includegraphics[width=1\textwidth]{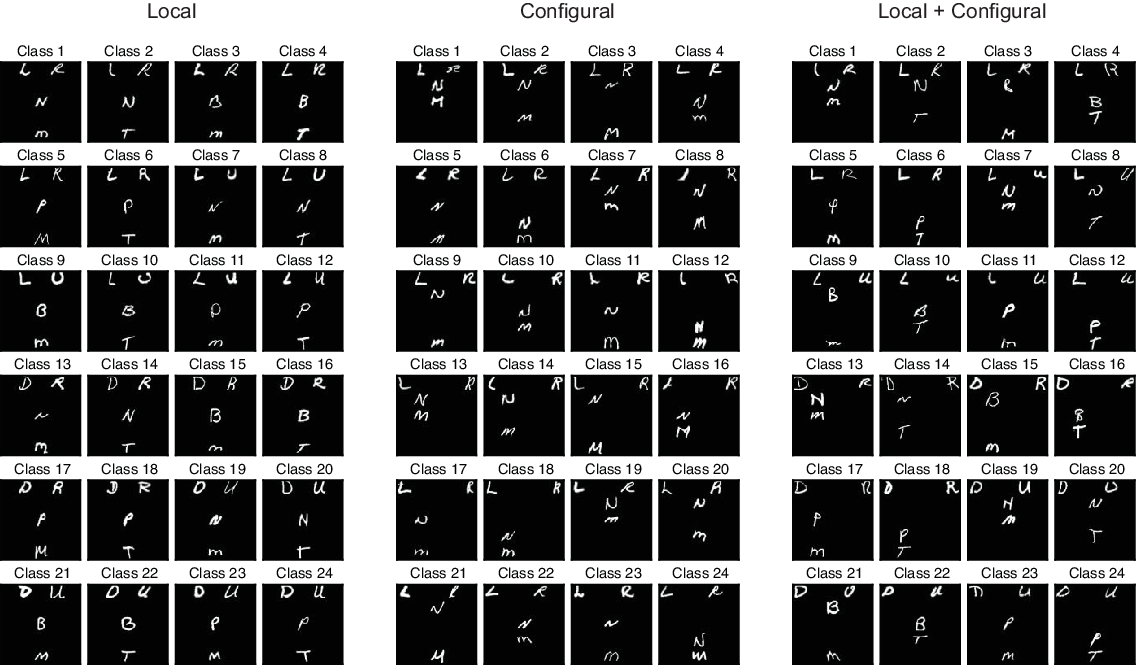}
\caption{Sample images from the total 24 classes for different tasks, with local (left), configural (middle), and local plus configural (right) tasks.}\label{sfig1}
\end{figure}

\begin{figure}[h]
\centering
\includegraphics[width=1\textwidth]{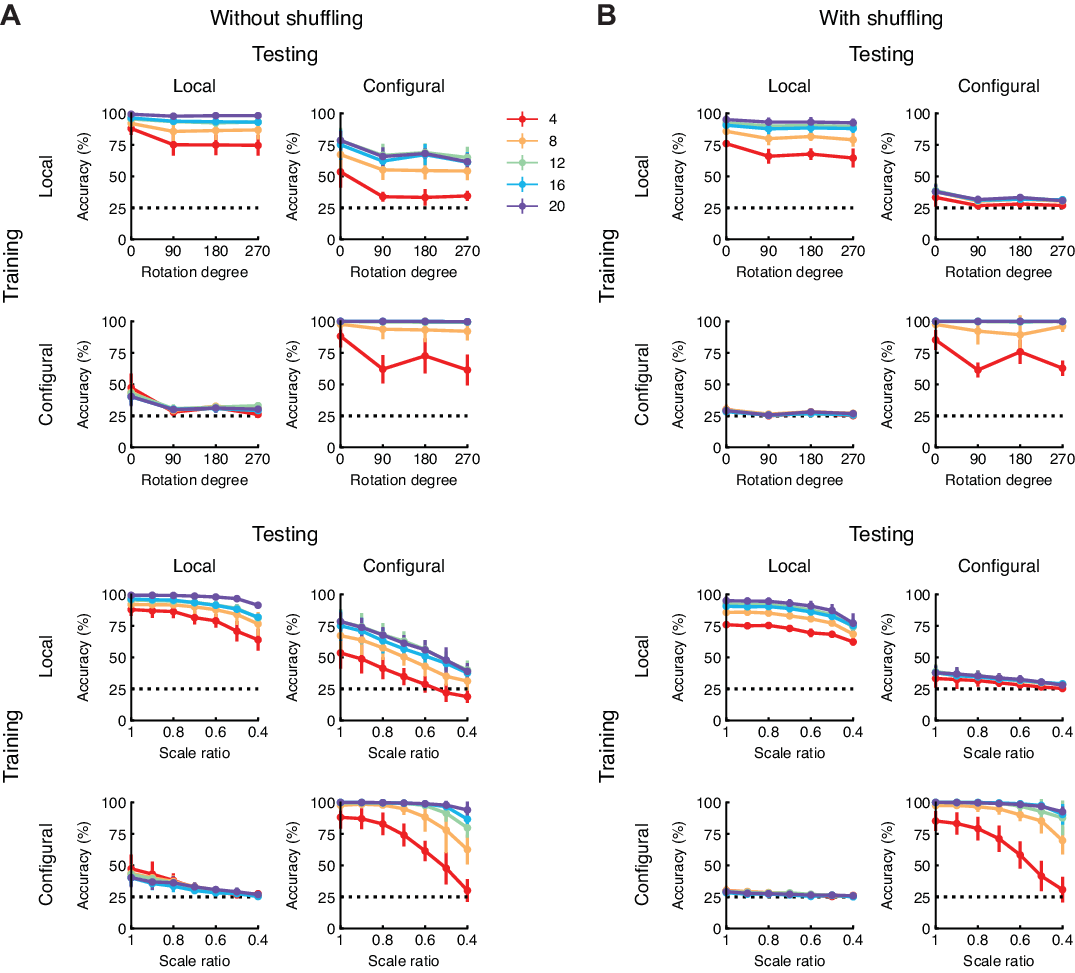}
\caption{Evaluation of network generalization performance across local and configural tasks, following the figure format used in \textbf{Fig. 1}. Performance was compared under scenarios where local feature elements were consistent (\textbf{A}) and randomly shuffled (\textbf{B}) during both training and evaluation phases.}\label{sfig2}
\end{figure}

\begin{figure}[h]
\centering
\includegraphics[width=0.6\textwidth]{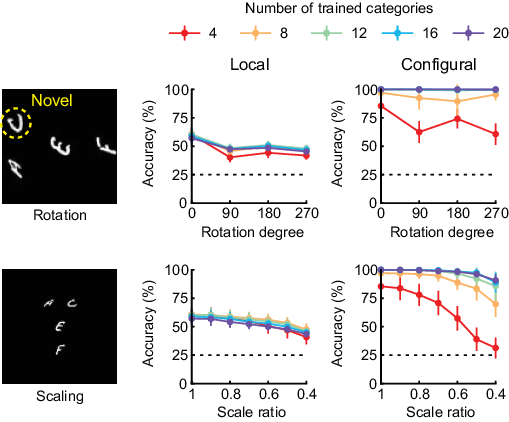}
\caption{Performance accuracy for local and configural tasks under rotation (top) and scaling (bottom) transformations, with patterns that included novel local features and had their local feature positions randomly shuffled. Different colors represent the number of categories trained.}\label{sfig3}
\end{figure}

\begin{figure}[h]
\centering
\includegraphics[width=1\textwidth]{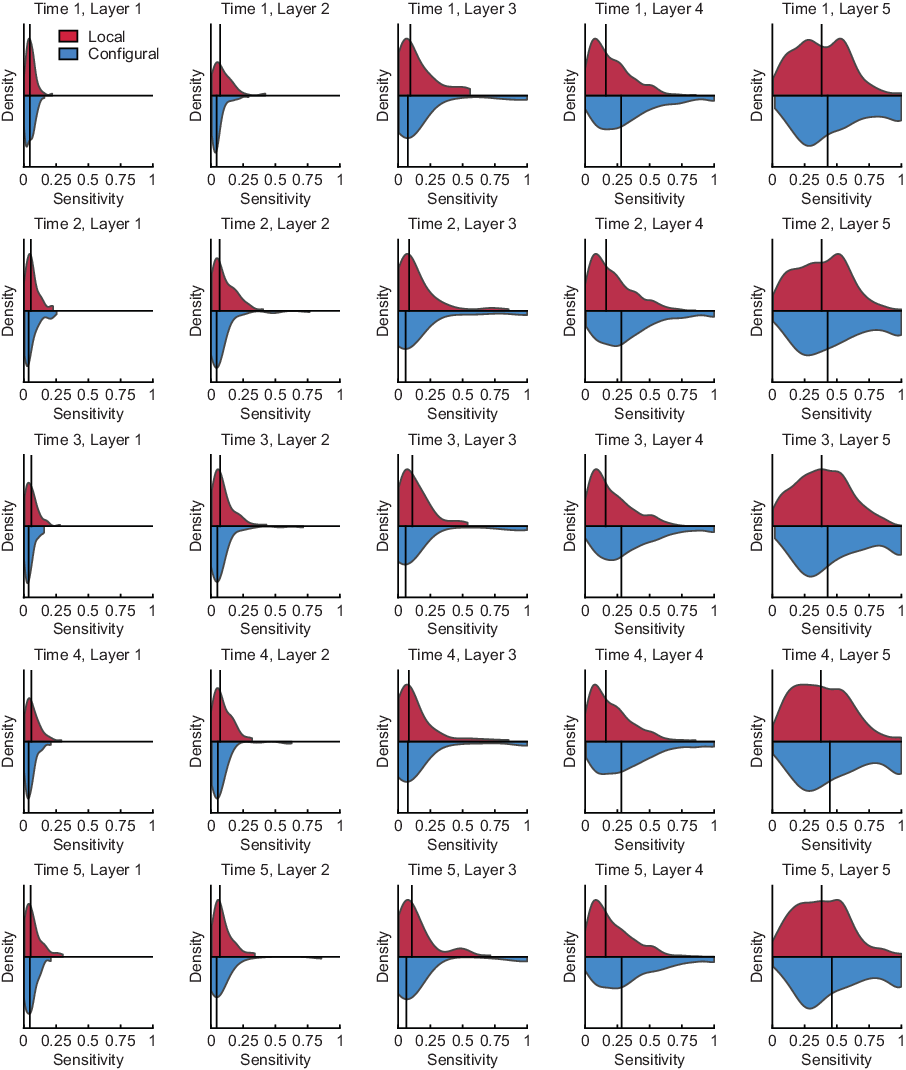}
\caption{Histograms displaying the sensitivity of individual neurons to local (red) and configural (blue) cues across the layers of ConvLSTM and across recurrent cycles.}\label{sfig4}
\end{figure}

\end{appendices}

\end{document}